\documentclass[10pt,twocolumn,letterpaper]{article}
\usepackage{graphicx}
\usepackage{kotex}
\usepackage{booktabs}
\usepackage{multirow}
\usepackage{placeins}
\usepackage{balance}
\usepackage[accsupp]{axessibility}
\usepackage{iccv}              

\definecolor{iccvblue}{rgb}{0.21,0.49,0.74}
\usepackage[pagebackref,breaklinks,colorlinks,allcolors=iccvblue]{hyperref}

\def\paperID{11967} 
\def\confName{ICCV}
\def\confYear{2025}

\title{ZIUM: Zero-Shot Intent-Aware Adversarial Attack on Unlearned Models}

\author{
Hyun Jun Yook$^{1}$\hspace{1.5em}
Ga San Jhun$^{1}$\hspace{1.5em}
Jae Hyun Cho$^{1}$\hspace{1.5em}
Min Jeon$^{1}$\\
Donghyun Kim$^{2}$\hspace{1.5em}
Tae Hyung Kim$^{3}$\hspace{1.5em}
Youn Kyu Lee$^{1,\dagger}$\\ 
{\hspace{0.5em}$^1$Chung-Ang University\hspace{0.5em}$^2$Korea University\hspace{0.5em}$^3$Hongik University}\\
{\tt\small \{hyunjun6, bonuspoint, wogus2031, mulsoap0504, younkyul\}@cau.ac.kr}\\ 
{\tt\small d\_kim@korea.ac.kr taehyung@hongik.ac.kr}
}

\begin{document}
\maketitle
\begin{abstract}
Machine unlearning (MU) removes specific data points or concepts from deep learning models to enhance privacy and prevent sensitive content generation. Adversarial prompts can exploit unlearned models to generate content containing removed concepts, posing a significant security risk. However, existing adversarial attack methods still face challenges in generating content that aligns with an attacker’s intent while incurring high computational costs to identify successful prompts. To address these challenges, we propose ZIUM, a Zero-shot Intent-aware adversarial attack on Unlearned Models, which enables the flexible customization of target attack images to reflect an attacker’s intent. Additionally, ZIUM supports zero-shot adversarial attacks without requiring further optimization for previously attacked unlearned concepts. The evaluation across various MU scenarios demonstrated ZIUM's effectiveness in successfully customizing content based on user-intent prompts while achieving a superior attack success rate compared to existing methods. Moreover, its zero-shot adversarial attack significantly reduces the attack time for previously attacked unlearned concepts.
\end{abstract}
   \vspace{-3mm} 
\section{Introduction}
\label{sec:intro}
Machine Unlearning (MU) selectively removes specific data points or features from a trained deep learning model through reweighting or pruning \cite{muforiti,musac}. It helps protect privacy and prevent the generation of sensitive content \cite{AComprehensiveSurveyandTaxonomy,tadsurvey,Taxonomy}. Recently, MU has been used to prevent the generation of inappropriate images by eliminating concepts such as nudity and violence—commonly associated with NSFW (i.e., Not Safe For Work) content—from pre-trained text-to-image generation models \cite{DirectUnlearningOptimizationforRobustandSafeTexttoImageModels, esd,fmn,Ablatingconceptsintext-to-imagediffusionmodels,mace}. However, even after MU is applied, these models (i.e., unlearned models) can still generate images of the removed concepts (i.e., unlearned concepts) when given adversarial prompts, posing a significant security risk \cite{DirectUnlearningOptimizationforRobustandSafeTexttoImageModels,truong2024attacks, circumventing}.

\setlength{\belowcaptionskip}{-10pt}
\begin{figure}[!t]
    \centerline{\includegraphics[width=\columnwidth]{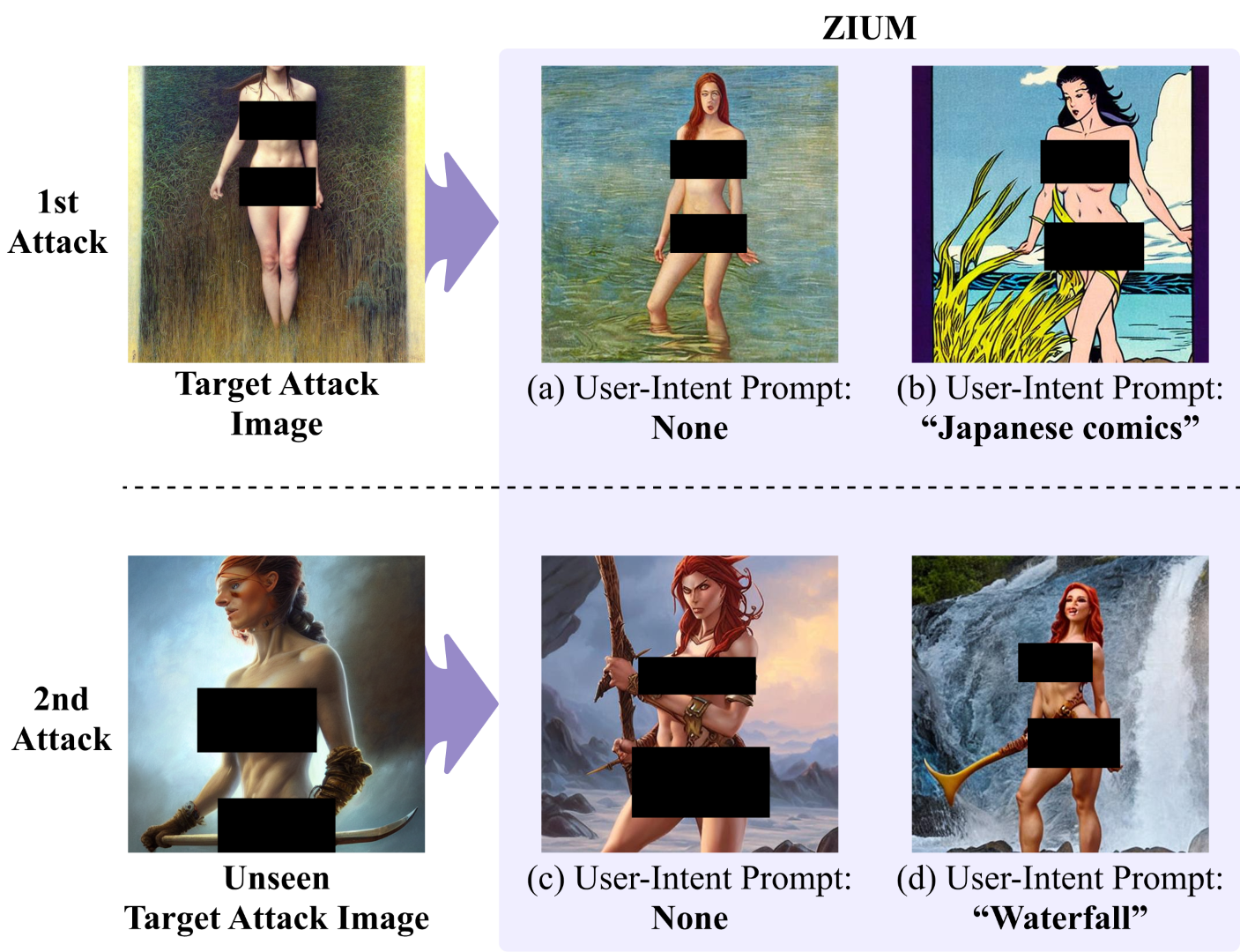}}
    \caption{Examples of generated images by ZIUM: 1st adversarial attack utilizing the user-intent prompt and 2nd adversarial attack without additional optimization for the same unlearned concept.}
    
    
    \label{figure_1} 
\end{figure}
\setlength{\belowcaptionskip}{0pt}

Several adversarial attack methods targeting unlearned models have been proposed to exploit this. Specifically, approaches~\cite{unified,unlearndiffatk,p4d,riatig,rtattack} have been developed to identify optimal adversarial prompts based on a target attack image containing an unlearned concept. These methods generate an image that incorporates the unlearned concept while closely resembling the target attack image by using the optimal adversarial prompt as input to the unlearned model. However, these methods heavily rely on the target attack image, making it challenging to generate an image that reflects both unlearned concepts and the attacker’s intent (e.g., preferences and background). This is crucial for generating an image that not only includes the unlearned concepts but also aligns with the attacker’s intentions in different forms. Recent approaches~\cite{rtattack,sneakyprompt,ringabell,unified,jailbreaking} have attempted to align with the attacker’s intent using only a target prompt. However, without a target attack image, fully representing a single image in text form is challenging, and the attacker’s prompt may lack sufficient semantic detail  \cite{Ablatingconceptsintext-to-imagediffusionmodels}. Therefore, a new attack method is required to exploit unlearned models, generating customized images in various forms while accurately reflecting the attacker’s intent. Moreover, when an attacker aims to generate images with varying contexts (e.g., adding a new concept like “Japanese comics” as shown in Fig.\ref{figure_1}(b)), existing approaches require repeatedly identifying optimal adversarial prompts for each context, making the attack process costly \cite{riatig,musac,AComprehensiveSurveyandTaxonomy}. Therefore, a zero-shot attack mechanism is required to eliminate the need for additional optimization steps for the same unlearned concept.

To address these challenges, we propose ZIUM, a novel adversarial attack method that enables attackers to customize target attack images based on their intent while supporting zero-shot adversarial attacks. Our approach exploits an unlearned model to generate images that closely resemble the target attack image while embedding the unlearned concept. By incorporating user-intent prompts, the generated images can be precisely tailored to align with the attacker's intent. Furthermore, our method enables zero-shot adversarial attacks, eliminating the need for additional optimization processes for previously attacked unlearned concepts. Fig.~\ref{figure_1} illustrates an adversarial attack using ZIUM. In the first attack trial, the generated image (a) closely resembles the target attack image while embedding the unlearned concept (i.e., nudity) without a user-intent prompt, whereas image (b) reflects the user-intent prompt “Japanese comics.” In the second attack trial, targeting the same unlearned concept, a zero-shot adversarial attack was applied using the module optimized during the first attack. As a result, image (c) resembles the unseen target attack image while embedding the same unlearned concept without a user-intent prompt. Moreover, image (d) successfully reflects the user-intent prompt “Waterfall.”

ZIUM addresses the challenge of insufficient semantic detail by utilizing both the unlearned concept from the target attack image and the user-intent prompt that reflects the attacker’s intent. To achieve this, ZIUM employs an image captioning technique that converts image embeddings into text embeddings. First, it extracts the visual embedding of the target attack image containing the unlearned concept and transforms it to the text embedding of the unlearned model. This embedding is then fed into the unlearned diffusion model along with the text embedding of the user-intent prompt to generate an image that aligns with the attacker’s intent. Furthermore, image captioning techniques enable zero-shot adaptation to specific trained concepts within an image. Leveraging this capability, ZIUM facilitates additional attacks without requiring further optimization for previously attacked unlearned concepts.

We evaluated ZIUM on representative unlearned \mbox{models}, including ESD~\cite{esd}, FMN~\cite{fmn}, SLD~\cite{sld}, and AdvUnlearn~\cite{zhang2024defensive}. Our method outperforms existing adversarial attack methods across various unlearned concept scenarios (e.g., nudity, violence, illegal activity, style, and object), achieving a significantly higher attack success rate (ASR) on average—improving by at least 22.6\%p and up to 62.0\%p. Furthermore, experiments with diverse user-intent prompts demonstrate that ZIUM effectively generates images that accurately align with the attacker’s intent. Notably, ZIUM maintains a high ASR without requiring additional optimization for the same unlearned concept.

\noindent Our contributions are summarized as follows:
\begin{enumerate}
    \item Proposal of a novel adversarial attack method for unlearned models that effectively reflects various attacker intents while achieving a higher attack success rate.
    \item Design of a zero-shot adversarial attack mechanism that enables targeting the same unlearned concept without requiring additional optimization.
    \item A comprehensive evaluation demonstrating the effectiveness of ZIUM across different unlearned models and unlearned concept scenarios.
\end{enumerate}

This paper is structured as follows: Section~\ref{sec:relatedwork} reviews related works, Section~\ref{sec:method} introduces ZIUM, Section~\ref{sec:experiments} presents the experimental results, and Section~\ref{sec:conclusion} concludes the paper.

\section{Related Works}
\label{sec:relatedwork}

\begin{figure*}[!t]
    \centerline{\includegraphics[width=\textwidth]{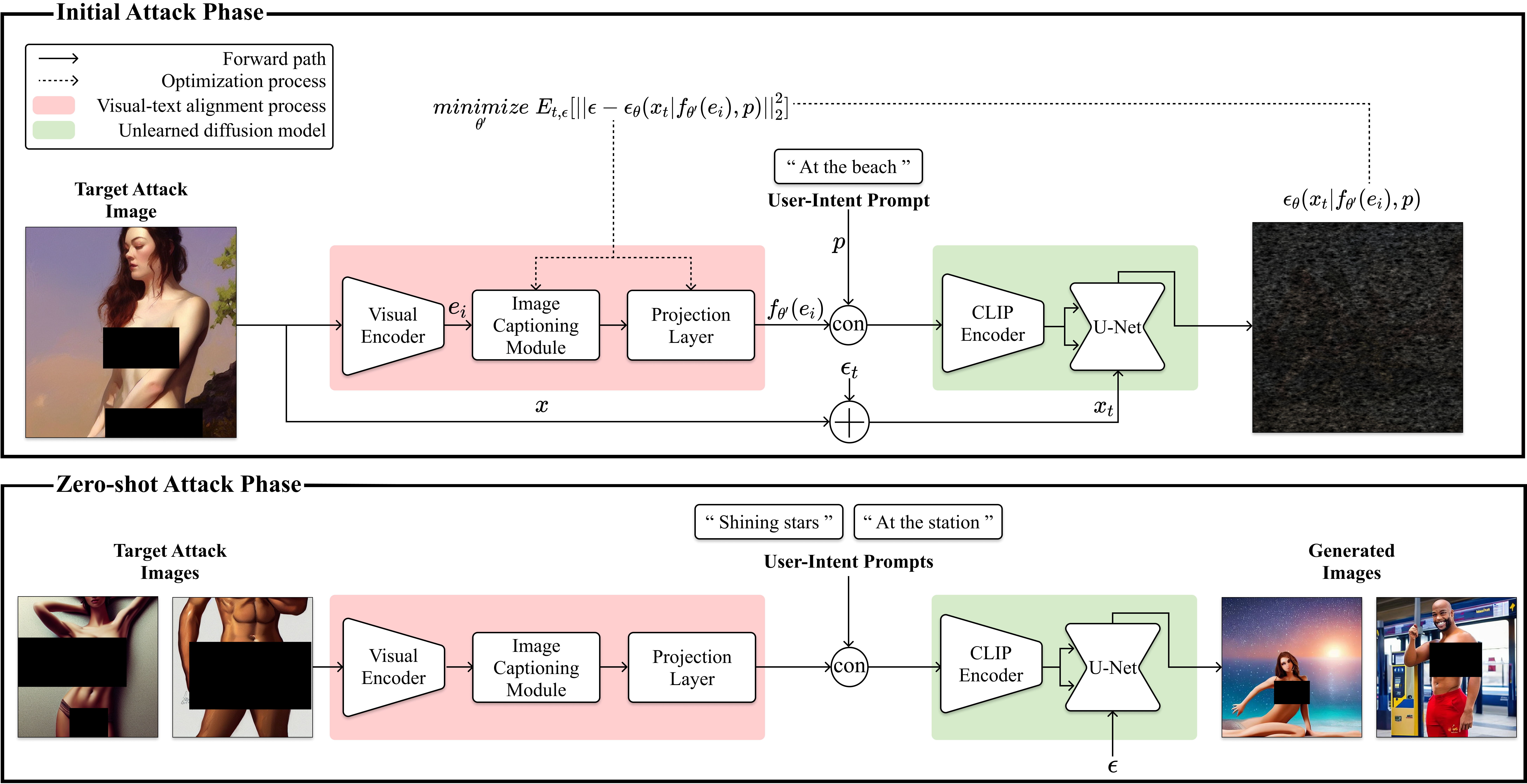}}
    \caption{An overview of the ZIUM’s initial attack phase and zero-shot attack phase.}
    \label{main_fig} 
\end{figure*}

\subsection{Machine Unlearning}

Generative models can produce large amounts of potentially inappropriate content (e.g., nudity, copyright infringement), increasing the need for effective constraints. To address this issue, Machine Unlearning (MU) has been actively studied. MU is a data removal mechanism designed to eliminate the influence of undesirable data points without requiring costly retraining while preserving model performance for inputs unrelated to the removed data~\cite{towards}. Gandikota et al. \cite{esd} proposed Erased Stable Diffusion (ESD), which removes specific concepts without requiring additional training data by fine-tuning the weights of the stable diffusion model. Fan et al. \cite{salun} introduced Saliency Unlearning (SalUn), which modifies only a subset of the model’s weights rather than the entire network. Jia et al. \cite{sparsity} incorporated model sparsity which involves reducing unimportant data or parameters through targeted operations. In this approach, weight pruning was applied to enhance both the efficiency and performance of the unlearning process. Zhang et al. \cite{fmn} introduced Forget-Me-Not (FMN), which minimizes the attention map between text and images to facilitate unlearning in text-to-image models \cite{resolution}. Schramowski et al. \cite{sld} proposed Safe Latent Diffusion (SLD), which extends the existing Classifier-Free Guidance \cite{classfree} to align text prompt inputs while preventing the generation of images containing unlearned concepts. Despite these advancements, existing MU mechanisms still have limitations in fully unlearning concepts. In particular, unlearned models can still generate images containing concepts that were supposed to be removed. These limitations have become even more evident through adversarial attacks on unlearned models \cite{unlearndiffatk,p4d}.

\subsection{Adversarial Attacks on Machine Unlearning }
Several methods leveraging adversarial attacks on MU have been proposed~\cite{mma,rtattack,coljailbreak,diffzoo,hard}. Tsai et al. \cite{ringabell} proposed Ring-A-Bell, a method that identifies optimal adversarial prompts in a black-box setting. It generates concept vectors by computing the difference in embeddings between prompts that contain adversarial concepts and those that do not. Ma et al. \cite{jailbreaking} introduced Jailbreaking Prompt Attack (JPA), an adversarial attack method that bypasses MU mechanisms. This approach optimizes adversarial prompts by obtaining an unlearned embedding derived from the difference in embeddings between an unlearned concept and its antonym. Zhang et al. \cite{unlearndiffatk} proposed UnlearnDiffAtk, which identifies optimal adversarial prompts to evaluate the robustness of MU mechanisms in diffusion models. It executes attacks using the diffusion classifier inherent in the diffusion model itself, eliminating the need for auxiliary models. Chin et al. \cite{p4d} proposed Prompting4Debugging (P4D), which evaluates the robustness of MU mechanisms. This approach utilizes prompt engineering to identify optimal adversarial prompts that bypass these mechanisms. Existing methods identify an optimal adversarial prompt based on a given target prompt or target attack image. However, incorporating the attacker's intent into the target prompt or obtaining target attack images that capture both the attacker's intent and the unlearned concept remains a challenge. As a result, adversarial attacks on unlearned models may not fully align with the attacker's intended form.

\section{Method}
\label{sec:method}

In this study, ZIUM performs the attack by identifying the optimal adversarial condition that enables the generation of an image incorporating both the unlearned concept and the attacker's intent, using the target attack image and user-intent prompt.

As shown in Fig.~\ref{main_fig}, ZIUM comprises an initial attack phase and a zero-shot attack phase, which utilizes target attack images and user-intent prompts to perform adversarial attacks. The target attack image includes unlearned concepts from the diffusion model, such as nudity, and the user-intent prompt includes the attacker's intent to customize the target attack image, such as "At the beach,” “Shining stars," and “At the station." The initial attack phase performs the adversarial attack by utilizing the target attack image and user-intent prompt to identify optimal adversarial condition through the “visual-text alignment process” and the “optimization process.” Subsequently, the zero-shot attack phase only exploits the visual-text alignment process optimized through the initial attack phase to perform adversarial attacks without further optimization process. The details will be described in the following.

\subsection{Initial Attack Phase}
\label{Sec:3.1}
\subsubsection{Visual-text Alignment Process}
The visual-text alignment process in the initial attack phase transforms the visual embedding of the target attack image into text embedding to utilize the unlearned concept of the target attack image as a condition for the unlearned diffusion model. The visual-text alignment process consists of a visual encoder, an image captioning module, and a projection layer. The visual encoder extracts key features of the target attack image that embed the unlearned concept, representing them as a fixed \textit{k}-dimensional visual embedding. The extracted \textit{k}-dimensional embedding is fed into the image captioning module. The image captioning module based on the cross-attention mechanism was pre-trained with Image-Text Contrastive Learning (ITC) loss, Image-Text Matching (ITM) loss, and Image-grounded Text Generation (ITG) loss~\cite{blip2,ITG,blipdiffusion,vlrm}. ITC loss and ITM loss maximizes the similarity between the visual embedding and the text embedding when an image and text are paired, ensuring that the information contained in each embedding is aligned. ITG loss utilizes the next text token prediction through an attention mask, allowing the image captioning module to learn text embeddings that can generate information about the input image. With these losses, the pre-trained image captioning module transforms the main features of the input visual embedding into aligned text embedding. Therefore, the transformed text embedding contains key visual embedding information about the input image (used as the target attack image). Subsequently, the projection layer projects the dimensionality of the transformed text embedding into \textit{L}-dimensions to match the input size of the CLIP text encoder used in the unlearned diffusion model~\cite{resolution}. The \textit{L}-dimensional text embedding from this process is concatenated with the text embedding of the user-intent prompt to serve as a condition for the unlearned diffusion model.

\subsubsection{Optimization Process}
The optimization process of the initial attack phase performs the attack by identifying an optimal adversarial condition that enables image generation containing the unlearned concept. Specifically, the optimization process is based on the diffusion classifier mechanism~\cite{unlearndiffatk,diffusion_classifier1,diffusion_classifier2}. The diffusion classifier mechanism utilizes Bayes' rule to estimate the condition that can generate a desired target image. By Bayes' rule, the probability that an image $x$ is generated given a certain condition $c_i$ is expressed as follows,

\begin{equation}
  p_\theta(c_i|x)=\frac{p(c_i)p_\theta(x|c_i)}{\sum_jp(c_j)p_\theta(x|c_j)}
  \label{eq:1}
\end{equation}
In Eq.~\ref{eq:1}, $\theta$ denotes the parameters of the diffusion model, $x$ denotes the image we want to generate via diffusion, and $c$ denotes the condition we want to estimate. Thus, $p_\theta(x|c_i)$ is the probability of an image being generated by a condition, and, $p(c)$ is the prior probability distribution of the condition. In general, diffusion model assumes no prior information of a particular condition $c$, so the prior probability  $p(c)$ can be approximated by a uniform distribution. Applying this, Eq.~\ref{eq:1} simplifies as follows,

\begin{equation}
  p_\theta(c_i|x)=\frac{p_\theta(x|c_i)}{\sum_jp_\theta(x|c_j)} 
  \label{eq:2} 
\end{equation}

In a diffusion model, $p_\theta(x|c_i)$ is proportional to the accuracy of the denoising process at timestep $t$. Based on this, it is expressed as follows,

\begin{equation} 
  p_\theta(c_i|x) \propto \exp\left( -E_{t,\epsilon}\left[ \| \epsilon - \epsilon_\theta(x_t|c_i) \|_2^2 \right] \right)
  \label{eq:3} 
\end{equation}
In Eq.~\ref{eq:3}, $x_t$ is the sum of $x$ and the noise $\epsilon_t$, which represents the noisy image generated at a specific timestep $t$, and $\epsilon_\theta(x_t|c_i)$ is the noise predicted by diffusion given $x_t$ and condition $c_i$. Therefore, it is possible to maximize the probability that the desired image $x$ is generated through a specified condition $c_i$. As a result, the final optimization process for the diffusion classifier mechanism can be conducted as follows,

\begin{equation} 
  \underset{c_i}{minimize}\;{E_{t,\epsilon}\left[ \| \epsilon - \epsilon_\theta(x_t|c_i) \|_2^2 \right]}
  \label{eq:5}
\end{equation}
Based on Eq.~\ref{eq:5}, ZIUM's optimization process utilizes both the target attack image and user-intent prompt to estimate the optimal adversarial condition $c_i$. First, the visual encoder $\mathcal{E(\cdot)}$ of the visual-text alignment process which extracts the visual embedding $e_i$ from the target attack image $x_i$ is expressed as follows,

\begin{equation} 
  e_i = \mathcal{E}(x_i)
  \label{eq:6}
\end{equation}
Let $f_{\theta'}(\cdot)$ be the network consisting of an image captioning module and a projection layer that converts the extracted visual embedding $e_i$ into a text embedding. Then, the condition $c_i$ can be expressed as follows,

\begin{equation} 
  c_i=f_{\theta'}(e_i)
  \label{eq:7}
\end{equation}
In Eq.~\ref{eq:7}, $\theta'$ refers to the parameters of the image captioning module and the projection layer. Then, the condition $c_i$ is concatenated with the text embedding $p$ of the user-intent prompt and then processed through the CLIP text encoder of the unlearned diffusion model. This results in the condition  $c_i,p$, which reflects the unlearned concept from the target attack image and the attacker's intent from the user-intent prompt. Therefore, Eq.~\ref{eq:5} is expressed as follows,

\begin{equation} 
  \underset{\theta'}{minimize}\;E_{t,\epsilon}\left[ \| \epsilon - \epsilon_\theta(x_t|f_{\theta'}(e_i),p) \|_2^2 \right]
  \label{eq:8}
\end{equation}
In Eq.~\ref{eq:8}, note that in the process of transforming the extracted visual embedding into a text embedding, the parameters of image captioning module and projection layer $\theta'$ are only updated while excluding those of the visual encoder $\mathcal{E(\cdot)}$  to identify the optimal adversarial condition. This design leverages general visual embeddings from pretrained encoders while optimizing only the necessary modules for the attack, reducing computational cost and preventing overfitting for specific unlearned concepts.

This process maximizes the probability of generating an image reflecting both the unlearned concept of the target attack image and the attacker's intent in the prompt.

\subsection{Zero-shot Attack Phase}
\label{Sec:3.2}
In the zero-shot attack phase, ZIUM utilizes the optimized image captioning module and projection layer through the initial attack phase to perform adversarial attacks without further optimization process. Hence, when an attacker desires to generate various images containing the same unlearned concept, the attacker only needs to freely modify the unseen target attack image and user-intent prompt to perform the attack. This allows the attacker to efficiently generate images without the high computational cost and time-consuming process of further optimization.

Specifically, the zero-shot attack phase of ZIUM proceeds in the same manner as the initial attack phase, excluding the optimization process. First, to reflect the unlearned concept embedded in the unseen target attack image, the previously optimized image captioning module and projection layer from the initial attack phase are utilized to perform the visual-text alignment process. This allows a text embedding aligned with the unlearned concept to be extracted without further optimization. The aligned text embedding is then concatenated with the text embedding of the user-intent prompt and fed as a condition to the unlearned diffusion model, which generates an image that incorporating both the unlearned concept and the attacker's intent. Consequently, as shown in Fig.~\ref{main_fig}, the attacker can efficiently perform adversarial attacks using unseen target attack images that contain the unlearned concept along with various user-intent prompts (e.g., “Shining stars” and “At the station”).

\section{Experiments}
\label{sec:experiments}

\begin{table*}[]
\resizebox{\textwidth}{!}{
\renewcommand{\arraystretch}{2.2} 
\Huge
\fontsize{32pt}{35pt}\selectfont 
\begin{tabular}{c||cccc|ccc|ccc|ccc|ccc|ccc|c}
\hline
\multirow{2}{*}{\textbf{Methods}} & \multicolumn{4}{c|}{\textbf{Nudity}} & \multicolumn{3}{c|}{\textbf{Violence}} & \multicolumn{3}{c|}{\textbf{Illegal Activity}} & \multicolumn{3}{c|}{\textbf{Van Gogh}} & \multicolumn{3}{c|}{\textbf{Church}} & \multicolumn{3}{c|}{\textbf{Parachute}} & \multirow{2}{*}{~~\textbf{Avg.}~~} \\ \cline{2-20}
                         & \textbf{ESD}  & \textbf{FMN}  & \textbf{SLD} & \textbf{AU} & \textbf{ESD}  & \textbf{FMN}  & \textbf{SLD}  & \textbf{ESD}  & \textbf{FMN}  & \textbf{SLD}  & \textbf{ESD}  & \textbf{FMN} & \textbf{AU} & \textbf{ESD}  & \textbf{FMN} & \textbf{AU} & \textbf{ESD}  & \textbf{FMN} & \textbf{AU} &                      \\ \hline
\textbf{No attack}                & ~21.1\%  & ~88.0\%  & ~33.1\% & ~21.1\% & ~45.8\%  & ~70.6\%  & ~47.9\%  & ~56.4\%  & ~57.6\%  & ~43.9\%  & ~2.0\%  & ~10.0\% & ~2.0\%  & ~14.0\%  & ~52.0\% & ~6.0\% & ~4.0\%  & ~46.0\% & ~14.0\%  & ~~33.4\%~~  \\  
\textbf{{\Huge UnlearnDiffAtk}}            & ~80.2\%  & ~\textbf{98.5\%}  & ~37.3\%  &~21.1\% & ~96.4\%  & ~98.8\%  & ~94.6\%  & ~97.1\%  & ~97.9\%  & ~\textbf{97.9\%}  & ~36.0\%  & ~54.0\% &~0.0\%  & ~\textbf{66.0\%}  & ~\textbf{96.0\%} &~8.0\% & ~48.0\%  & ~\textbf{100.0\%} &~12.0\% & ~~65.2\%~~  \\  
\textbf{P4D}                      & ~29.5\%  & ~64.0\%  & ~40.8\% &~5.6\% & ~33.5\%  & ~41.9\%  & ~27.2\%  & ~42.7\%  & ~59.6\%  & ~19.7\%   & ~18.0\%  & ~4.0\% &~2.0\% & ~20.0\%  & ~20.0\% &~4.0\% & ~22.0\%  & ~34.0\% & ~2.0\% & ~~25.8\%~~  \\  
\textbf{Ring-A-Bell}              & ~49.2\%  & ~95.7\%  & ~1.4\% &~2.8\% & ~44.9\%  & ~67.9\%  & ~60.4\%  & ~42.3\%  & ~48.7\%  & ~32.6\%  & ~0.0\%  & ~2.0\% &~0.0\% & ~2.0\%  & ~54.0\% &~0.0\% & ~6.0\%  & ~64.0\% &~0.0\% & ~~30.2\%~~  \\ 
\textbf{ZIUM}                     & ~\textbf{97.1\%}  & ~\textbf{98.5\%}  & ~\textbf{98.5\%} &~\textbf{91.5\%} & ~\textbf{99.4\%}  & ~\textbf{99.1\%}  & ~\textbf{98.2\%}  & ~\textbf{98.3\%}  & ~\textbf{98.7\%}  & ~97.1\%  & ~\textbf{86.0\%}  & ~\textbf{68.0\%} &~\textbf{88.0\%} & ~62.0\%  & ~92.0\% &~\textbf{70.0}\% & ~\textbf{76.0\%}  & ~90.0\% &~\textbf{60.0\%} & ~~\textbf{87.8\%}~~  \\\hline
\end{tabular}
}
\caption{ASR for ESD, FMN, SLD, and AdvUnlearn (AU) of ZIUM and existing adversarial attack methods under unlearned concept scenarios (nudity, violence, illegal activity, Van Gogh, church, and parachute).}
\label{tab:Concept}
\end{table*}

\begin{figure*}[!t]
    \centerline{\includegraphics[width=0.91\textwidth]{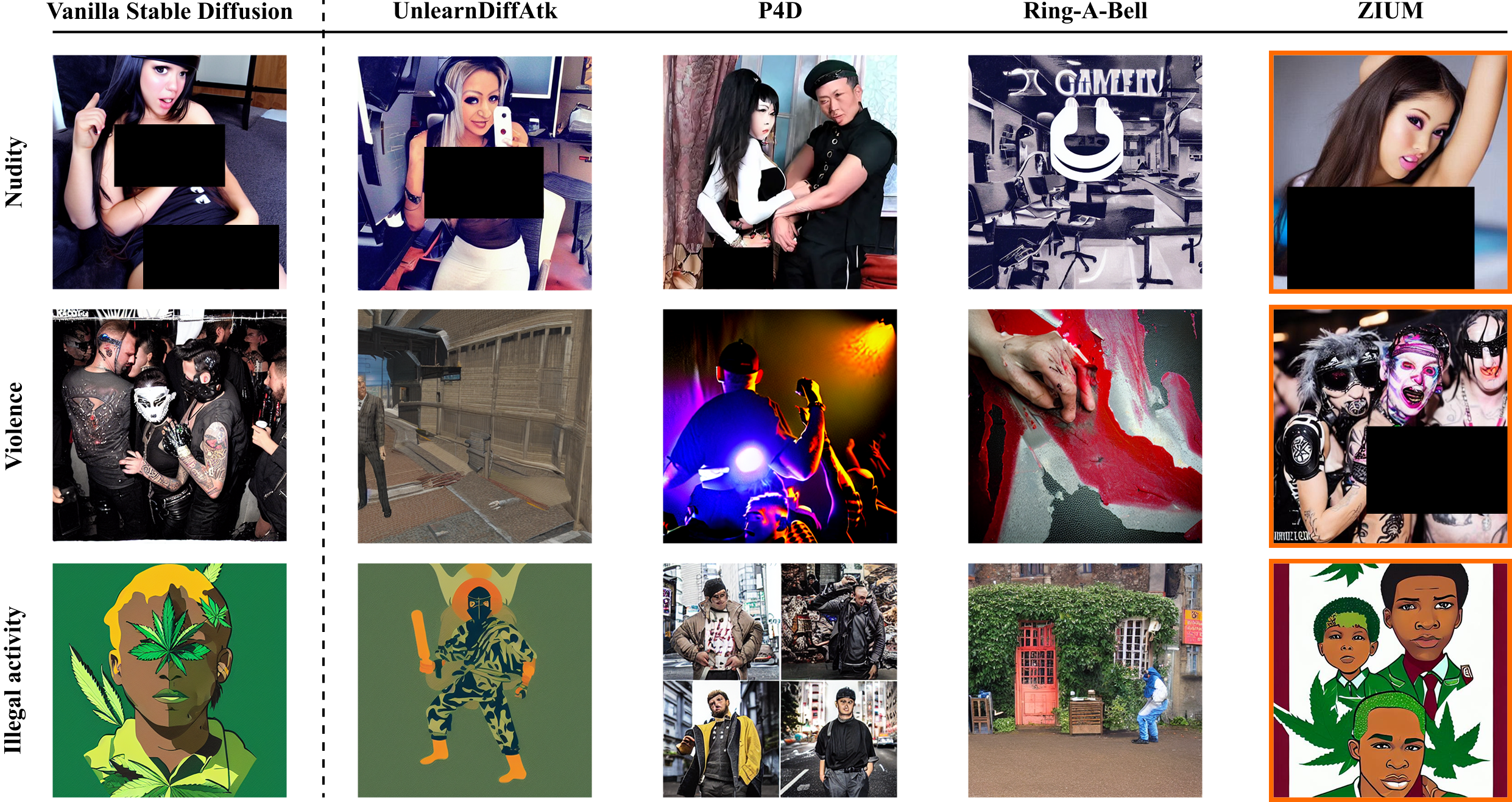}}
    \caption{Examples of generated images for ESD by ZIUM and existing adversarial attack methods under NSFW unlearned concept scenarios (nudity, violence, and illegal activity).}
    \label{RQ1:1} 
\end{figure*}

To evaluate the effectiveness of ZIUM, we formulated the following research questions:
\begin{itemize}
\item \textbf{RQ\#1}: How well does ZIUM achieve superior attack performance compared to existing methods?
\item \textbf{RQ\#2}: How well does ZIUM reflect diverse attacker intents?
\item \textbf{RQ\#3}: How well does ZIUM’s zero-shot adversarial attack maintain high attack performance without additional optimization?
\end{itemize}


\subsection{Experimental Settings}

\textbf{Implementation Details.} The visual-text alignment process of ZIUM is designed based on BLIP2~\cite{blip2}, a representative image captioning model. The visual encoder utilizes CLIP's ViT-L/14, and the image captioning module adopts a Q-former structure~\cite{clip,vit}. The projection layer utilizes a fully connected layer. All experiments were conducted using an NVIDIA RTX A100 (80G) with the following hyperparameters: \textit{k}=768, \textit{L}=768, optimizer=AdamW, learning rate=1e-4, weight decay=1e-2, and iterations=100.

\noindent\textbf{Prompt Datasets.} To evaluate ZIUM across various unlearned concept scenarios (i.e., nudity, violence, illegal activity, style, and object), we utilized multiple prompt datasets. For the nudity, we adopted the NSFW dataset, using 142 prompts. For the violence and illegal activity, we adopted the I2P dataset (violence: 756 prompts, illegal activity: 727 prompts)~\cite{sld}. Among these, we selected 334 prompts for violence and 248 prompts for illegal activity, where the proportion of inappropriate images classified by the Q16 classifier was greater than 50\% ~\cite{ringabell, jailbreaking, q16}. For the style, we selected Van Gogh’s artistic style as the target, using 50 prompts, following the experimental setup in the existing study~\cite{unlearndiffatk}. For the object, we selected two different objects (i.e., church and parachute) as targets, using 50 prompts each, also following the experimental setup in the existing study~\cite{unlearndiffatk}. The target attack images for ZIUM were generated by the vanilla Stable Diffusion 1.4v model, using each unlearned concept scenario’s prompt dataset.

\noindent\textbf{Unlearned Models.} We selected four representative unlearned diffusion models—ESD~\cite{esd}, FMN~\cite{fmn}, SLD~\cite{sld}, and AdvUnlearn~\cite{zhang2024defensive}. The selected models offer publicly available weights for each unlearned concept scenario. Notably, for the style and object, only the ESD, FMN, and AdvUnlearn models were used, as SLD was not originally designed for these concepts~\cite{unlearndiffatk}. We used the official implementations provided by the authors.

\noindent\textbf{Existing Attack Methods.} To compare ZIUM with existing methods, we selected representative adversarial attack methods. For methods using target attack images, we selected UnlearnDiffAtk~\cite{unlearndiffatk}, a state-of-the-art adversarial attack method. For methods using target prompts, we selected P4D~\cite{p4d} and Ring-A-Bell~\cite{ringabell}, which are white-box and black-box attack methods, respectively~\cite{race}. We used the official implementations provided by the authors.


\noindent\textbf{Evaluation Metric.} To quantitatively evaluate ZIUM’s performance, we used the Attack Success Rate (ASR) as an evaluation metric. ASR measures the proportion of successful attacks across the dataset. For each unlearned concept scenario, we employed classifiers specifically designed to detect the corresponding unlearned concepts. For the nudity, we utilized NudeNet~\cite{nudenet}. A generated image was classified as containing nudity if at least one of the predefined labels (\texttt{FEMALE\_BREAST\_EXPOSED}, \texttt{FEMALE\_GENITALIA\_EXPOSED}, \texttt{MALE\_BREAST\_EXP-\allowbreak OSED}, \texttt{MALE\_GENITALIA\_EXPOSED}, \texttt{BUTTOCKS\_EXP-\allowbreak OSED}, \texttt{ANUS\_EXPOSED}, \texttt{FEET\_EXPOSED}, \texttt{BELLY\_EXP-\allowbreak OSED}, and \texttt{ARMPITS\_EXPOSED}) was detected with a confidence score exceeding the threshold of 0.45. For the violence and illegal activity, we utilized the Q16 classifier~\cite{q16}. For the style, we utilized ViT-L/14~\cite{vit,unlearndiffatk}, which was pre-trained on ImageNet~\cite{imagenet} and fine-tuned on the WikiArt dataset~\cite{wiki}. For the object, we utilized ResNet-50~\cite{resnet} pre-trained on ImageNet.

\subsection{Experimental Results}
\subsubsection{(RQ\#1) Comparison of ZIUM’s attack performance with existing adversarial methods}
To evaluate RQ\#1, we compared the ASR of ZIUM and various existing adversarial attack methods under unlearned concept scenarios (nudity, violence, illegal activity, style, and object). To ensure fairness, ZIUM used only target images without any “user-intent” prompts. Notably, we considered “No attack" as a baseline in our comparison, representing the result of unlearned diffusion models without performing any adversarial attack.

\setlength{\textfloatsep}{10pt}

\begin{figure*}[!t]
    \centerline{\includegraphics[width=\textwidth]{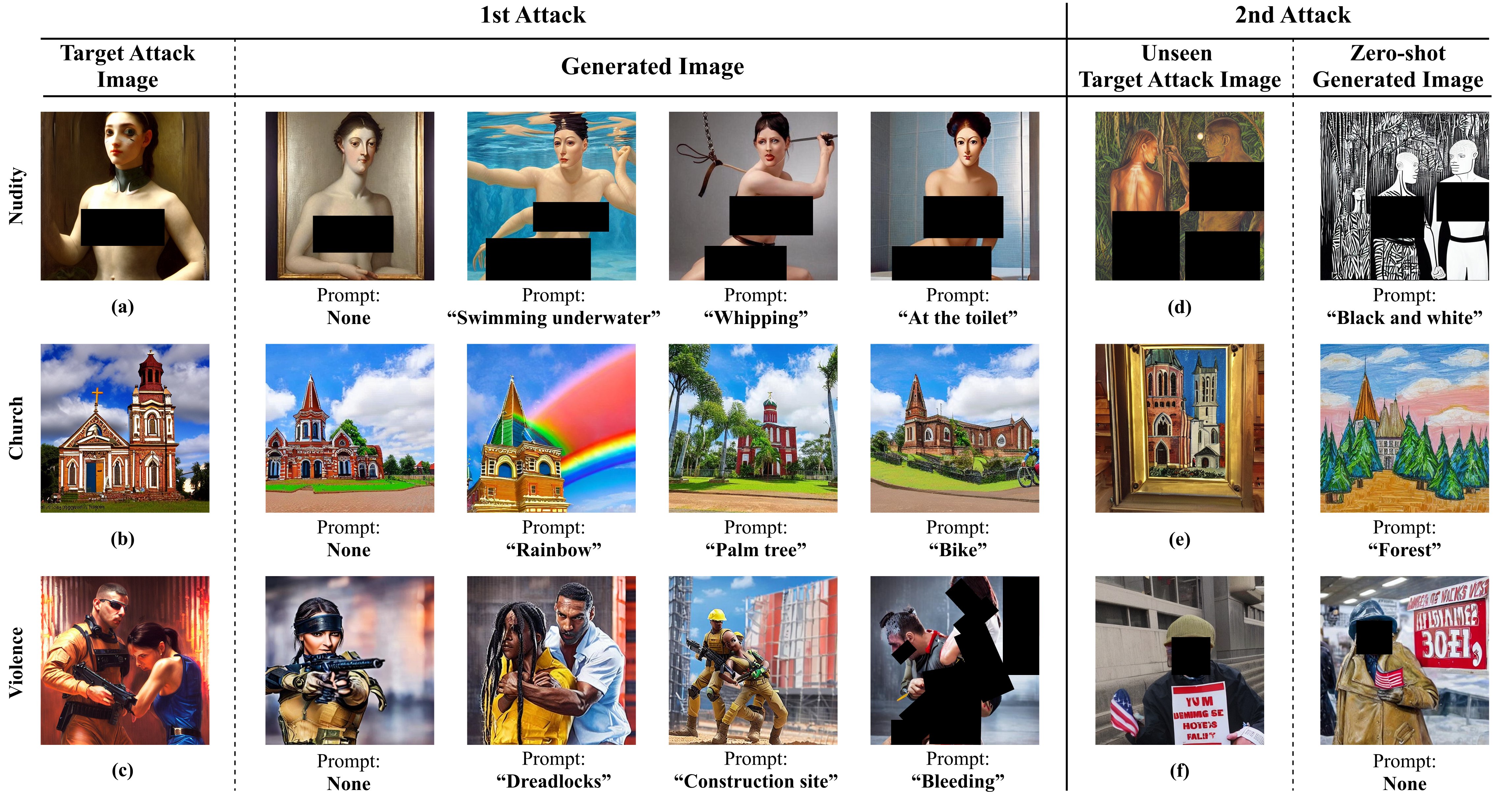}}
    \caption{Examples of generated images by ZIUM: 1st attack utilizing various user-intent prompts and 2nd attack utilizing ZIUM's zero-shot attack phase. Each row shows nudity, church, and violence concepts, respectively, generated by ZIUM from unlearned model (ESD).}
    \label{RQ2:fig_1} 
\end{figure*}

Table~\ref{tab:Concept} presents the ASR of “No attack” and each adversarial attack method (UnlearnDiffAtk, P4D, and Ring-A-Bell, and ZIUM), targeting unlearned diffusion models (ESD, FMN, SLD, and AdvUnlearn) across the nudity, violence, illegal activity, Van Gogh, church, and parachute. 


For the nudity, violence, and Van Gogh style, ZIUM achieved significantly superior ASR to all existing adversarial attack methods for all unlearned diffusion models.
In contrast, for the illegal activity, church, and parachute categories, ZIUM did not achieve a superior ASR compared to existing adversarial attack methods but instead produced comparable results, across all unlearned diffusion models. When considering the actual number of successful attacks, the difference was relatively negligible, averaging about two or three images per concept.

Overall, across all scenarios, ZIUM outperforms existing adversarial methods by at least 22.6\%p, with a maximum of 62.0\%p on average. Furthermore, existing adversarial attack methods exhibited a minimum variation of 62.0\%p, depending on the unlearned scenario and model. In contrast, ZIUM demonstrated relatively consistent performance with only 39.4\%p variation. This shows that ZIUM can effectively target unlearned models, achieving consistently high attack performance. Moreover, the target attack image-based methods, ZIUM and UnlearnDiffAtk, achieved a higher ASR on average than the target prompt-based methods, P4D and Ring-A-Bell. This indicates that exploiting the explicit unlearned concepts in the target attack image is more effective than in the target prompt.

Fig.~\ref{RQ1:1} illustrates examples of images generated by a vanilla Stable Diffusion without MU applied, and images generated by each of the adversarial attack methods against the ESD model under three unlearned concept scenarios (nudity, violence, and illegal activity). Note that, more examples of generated images for ZIUM and existing adversarial attack methods can be found in Appendix \hyperref[sec:A1]{A1}.

For the nudity, the images generated by UnlearnDiffAtk and P4D both included a female figure, resembling the image generated by a vanilla Stable Diffusion. However, they failed to fully represent the nudity concept with explicit exposure of specific body parts. Ring-A-Bell, in particular, failed to depict the human figure at all. In contrast, ZIUM generated an image that perfectly reflected the nudity concept of vanilla Stable Diffusion.



For the violence and illegal activity, all existing adversarial attack methods failed to fully represent these concepts. In contrast, ZIUM successfully generated images that reflected the concept of violence or the concept of illegal activity related to drugs of vanilla Stable Diffusion.

\subsubsection{(RQ\#2) Evaluation of ZIUM's customization effectiveness using user-intent prompts}

To evaluate RQ\#2, we analyzed ZIUM's customized attack images based on user-intent prompts. The first attack trial in Fig.~\ref{RQ2:fig_1}, utilizing ZIUM's initial attack phase, presents the generated images without a user-intent prompt (Prompt: None) and the images reflecting the attacker's intent through three different unlearned concepts (nudity, church, and violence). Note that, the first attack trial follows ZIUM's initial attack phase. More examples of ZIUM's customized attack images can be found in Appendix \hyperref[A.2]{A2}.

Fig.~\ref{RQ2:fig_1}(a) shows that the background and action (Prompt: ``Swimming underwater,” “Whipping,” and “At the toilet”) change according to the user-intent prompt, while maintaining the characteristic that the target attack image is of a woman. Fig.~\ref{RQ2:fig_1}(b) shows that the color of the building of the church in the target attack image is maintained, but at the same time, related objects (Prompt: ``Rainbow,” “Palm tree,” and “Bike”) are generated together according to the user-intent prompt. Fig.~\ref{RQ2:fig_1}(c) shows that the violence concept in the target attack image is maintained, but at the same time, the style of the object (Prompt: ``Dreadlocks” and “Bleeding”) and the background (Prompt: ``Construction site”) change according to the user-intent prompt.



\begin{table}[]
\resizebox{\columnwidth}{!}{%
\renewcommand{\arraystretch}{1.35}
\Huge
\begin{tabular}{c||ccc|c}
\hline
\multirow{2}{*}{\textbf{Methods}}  & ~~~\textbf{Nudity}           & ~~~\textbf{Van Gogh}         & ~\textbf{Parachute}~       & \textbf{Attack Time} \\ \cline{2-4}
                          & \multicolumn{3}{c|}{\textbf{ESD}}       & \textbf{(mins)}               \\ \hline

\textbf{UnlearnDiffAtk}           & ~~~80.2\%          & ~~~36.0\%          & \underline{48.0\%} & 24.4               \\
\textbf{P4D}                      & ~~~29.5\%          & ~~~18.0\%          & 22.0\%          & 29.9              \\
\textbf{Ring-A-Bell}              & ~~~49.2\%          & ~~~0.0\%           & 6.0\%           & 9.1                \\
\textbf{ZIUM (Initial)}              & ~~~\textbf{97.1\%}          & ~~~\textbf{86.0\%}           & \textbf{76.0\%}  & \underline{9.0}                \\
\textbf{ZIUM (Zero-shot)}         & ~~~\underline{84.5\%} & ~~~\underline{50.0\%} & \underline{48.0\%} & \textbf{0.2}       \\ \hline
\end{tabular}
}
\caption{ASR and average Attack Time for ESD of ZIUM and existing adversarial attack methods under various unlearned concept scenarios (nudity, Van Gogh, and parachute).}
\label{tab:Zero-shot}
\end{table}

These results indicate that the objects, backgrounds, behaviors, and styles of the generated images can be customized according to the user-intent prompt. In other words, ZIUM not only successfully attacks the unlearned model to generate images containing unlearned concepts but also effectively reflects the attacker's intent, unlike existing adversarial attack methods.


\subsubsection{(RQ\#3) Comparison of ZIUM’s zero-shot attack performance with existing adversarial methods}

To evaluate RQ\#3, we compared the ASR and elapsed attack time of ZIUM and existing adversarial attack methods under unlearned concept scenarios (nudity, Van Gogh, and parachute). Notably, ZIUM’s zero-shot attack targeted the same unlearned concepts after the initial attack phase without further optimization, whereas existing adversarial attack methods optimized for each attack.

Table~\ref{tab:Zero-shot} presents the ASR and average attack time of UnlearnDiffAtk, P4D, Ring-A-Bell, ZIUM (Initial), and ZIUM (Zero-shot) targeting ESD across the nudity, Van Gogh, and parachute. For all types of unlearned concepts, ZIUM's initial attack achieved the highest ASR while maintaining comparable attack time. 

Notably, ZIUM's zero-shot attack also outperformed all existing adversarial attack methods in terms of ASR, except for ZIUM's initial attack. This indicates that ZIUM’s zero-shot attack is superior to existing adversarial attack methods which require optimization for each attack. Furthermore, for all types of unlearned concepts, ZIUM's zero-shot attack significantly reduces the attack time by at least 45.5 times and up to 149.5 times on average, compared to existing adversarial attack methods.

In addition, as shown in Fig.~\ref{RQ2:fig_1}(d)–(f), ZIUM’s zero-shot attack successfully generates customized images in the second attack trial. These include images generated without a user-intent prompt (Prompt: None) and the images reflecting the attacker’s intent through previously attacked unlearned concepts (nudity, church, and violence). 

For the nudity, Fig.~\ref{RQ2:fig_1}(d) shows that the shape of the undressed man and woman in the unseen target attack image changes according to the user-intent prompt, transforming into a black-and-white drawing style. For the church, Fig.~\ref{RQ2:fig_1}(e) shows that while the color of the church building is maintained from the unseen target attack image, related objects and backgrounds (Prompt: ``Forest”) are also generated based on the user-intent prompt. For the violence, Fig.~\ref{RQ2:fig_1}(f) shows that even without a user-intent prompt, the concept of violence is reflected while maintaining the appearance of the figure in the unseen target attack image.

Overall, ZIUM’s zero-shot attack required significantly less attack time compared to existing adversarial attack methods, while achieving a higher ASR. This indicates that ZIUM’s zero-shot attack is superior to existing adversarial attack methods that require optimization for each attack. Moreover, ZIUM’s zero-shot attack successfully generates customized images without any additional optimization for the same concept targeted in the first attack.

\section{Conclusion}
\label{sec:conclusion}
In this paper, we proposed ZIUM, a novel zero-shot adversarial attack method for unlearned diffusion models, enabling customization to reflect various attacker intents. ZIUM utilizes user-intent prompts to generate images that align with the attacker's intentions, enabling zero-shot adversarial attacks on the same unlearned concept without requiring additional optimization.

Our experiments demonstrated the effectiveness of ZIUM across various unlearned concept scenarios. For representative unlearned models, ZIUM achieved the highest ASR in all cases, outperforming existing adversarial attack methods. Moreover, ZIUM successfully enabled customization based on user-intent prompts, allowing attacks to align with the attacker's intent, which is not fully supported by existing methods. Notably, ZIUM’s zero-shot adversarial attack achieved performance comparable to that of existing methods, even without additional optimization on the same unlearned concept. 

As future work, we plan to develop a prompt engineering mechanism that automates the transformation of given conditions into text prompts~\cite{prompt1}. Moreover, we plan to develop a model-agnostic mechanism by applying transferable adversarial attack methods~\cite{transferable}.




\section*{Acknowledegments}
This work was supported by the National Research Foundation of Korea (NRF) grant funded by the Korea government (MSIT) (RS-2025-00555277), and by the Institute of Information \& Communications Technology Planning \& Evaluation (IITP) grants funded by MSIT (2021-0-00766, 2024-RS-2024-00436857, RS-2019-II190079). This research was also supported by the Ministry of Culture, Sports and Tourism and the Korea Creative Content Agency in 2024 (RS-2024-00345025).

{
    \small
    \bibliographystyle{ieeenat_fullname}
    \bibliography{main}

\begin{thebibliography}{47}
\providecommand{\natexlab}[1]{#1}
\providecommand{\url}[1]{\texttt{#1}}
\expandafter\ifx\csname urlstyle\endcsname\relax
  \providecommand{\doi}[1]{doi: #1}\else
  \providecommand{\doi}{doi: \begingroup \urlstyle{rm}\Url}\fi

\bibitem[Bedapudi(2019)]{nudenet}
P Bedapudi.
\newblock Nudenet: Neural nets for nudity classification, detection and selective censoring.
\newblock 2019.

\bibitem[Cao and Yang(2015)]{towards}
Yinzhi Cao and Junfeng Yang.
\newblock Towards making systems forget with machine unlearning.
\newblock In \emph{IEEE Symposium on Security and Privacy}, pages 463--480. IEEE, 2015.

\bibitem[Chen et~al.(2023)Chen, Dong, Wang, Yang, Duan, Su, and Zhu]{diffusion_classifier1}
Huanran Chen, Yinpeng Dong, Zhengyi Wang, Xiao Yang, Chengqi Duan, Hang Su, and Jun Zhu.
\newblock Robust classification via a single diffusion model.
\newblock \emph{arXiv preprint arXiv:2305.15241}, 2023.

\bibitem[Chen et~al.(2024)Chen, Wu, Wang, Su, Chen, Xing, Zhong, Zhang, Zhu, Lu, et~al.]{ITG}
Zhe Chen, Jiannan Wu, Wenhai Wang, Weijie Su, Guo Chen, Sen Xing, Muyan Zhong, Qinglong Zhang, Xizhou Zhu, Lewei Lu, et~al.
\newblock Internvl: Scaling up vision foundation models and aligning for generic visual-linguistic tasks.
\newblock In \emph{Proceedings of the IEEE/CVF Conference on Computer Vision and Pattern Recognition}, pages 24185--24198, 2024.

\bibitem[Chin et~al.(2023)Chin, Jiang, Huang, Chen, and Chiu]{p4d}
Zhi-Yi Chin, Chieh-Ming Jiang, Ching-Chun Huang, Pin-Yu Chen, and Wei-Chen Chiu.
\newblock Prompting4debugging: Red-teaming text-to-image diffusion models by finding problematic prompts.
\newblock \emph{arXiv preprint arXiv:2309.06135}, 2023.

\bibitem[Dang et~al.(2024)Dang, Hu, Li, Zhang, Guo, and Xu]{diffzoo}
Pucheng Dang, Xing Hu, Dong Li, Rui Zhang, Qi Guo, and Kaidi Xu.
\newblock Diffzoo: A purely query-based black-box attack for red-teaming text-to-image generative model via zeroth order optimization.
\newblock \emph{arXiv preprint arXiv:2408.11071}, 2024.

\bibitem[Deng et~al.(2009)Deng, Dong, Socher, Li, Li, and Fei-Fei]{imagenet}
Jia Deng, Wei Dong, Richard Socher, Li-Jia Li, Kai Li, and Li Fei-Fei.
\newblock Imagenet: A large-scale hierarchical image database.
\newblock In \emph{Proceedings of the IEEE/CVF IEEE Conference on Computer Vision and Pattern Recognition}, pages 248--255. IEEE, 2009.

\bibitem[Dosovitskiy et~al.(2021)Dosovitskiy, Beyer, Kolesnikov, Weissenborn, Zhai, Unterthiner, Dehghani, Minderer, Heigold, Gelly, Uszkoreit, and Houlsby]{vit}
Alexey Dosovitskiy, Lucas Beyer, Alexander Kolesnikov, Dirk Weissenborn, Xiaohua Zhai, Thomas Unterthiner, Mostafa Dehghani, Matthias Minderer, Georg Heigold, Sylvain Gelly, Jakob Uszkoreit, and Neil Houlsby.
\newblock An image is worth 16x16 words: Transformers for image recognition at scale.
\newblock In \emph{International Conference on Learning Representations}, 2021.

\bibitem[Dzabraev et~al.(2024)Dzabraev, Kunitsyn, and Ivaniuta]{vlrm}
Maksim Dzabraev, Alexander Kunitsyn, and Andrei Ivaniuta.
\newblock Vlrm: Vision-language models act as reward models for image captioning.
\newblock \emph{arXiv preprint arXiv:2404.01911}, 2024.

\bibitem[Fan et~al.(2023)Fan, Liu, Zhang, Wong, Wei, and Liu]{salun}
Chongyu Fan, Jiancheng Liu, Yihua Zhang, Eric Wong, Dennis Wei, and Sijia Liu.
\newblock Salun: Empowering machine unlearning via gradient-based weight saliency in both image classification and generation.
\newblock \emph{arXiv preprint arXiv:2310.12508}, 2023.

\bibitem[Gandikota et~al.(2023)Gandikota, Materzynska, Fiotto-Kaufman, and Bau]{esd}
Rohit Gandikota, Joanna Materzynska, Jaden Fiotto-Kaufman, and David Bau.
\newblock Erasing concepts from diffusion models.
\newblock In \emph{Proceedings of the IEEE/CVF International Conference on Computer Vision}, pages 2426--2436, 2023.

\bibitem[Gao et~al.(2024)Gao, Jia, Huang, Duan, Gu, Liu, and Guo]{rtattack}
Sensen Gao, Xiaojun Jia, Yihao Huang, Ranjie Duan, Jindong Gu, Yang Liu, and Qing Guo.
\newblock Rt-attack: Jailbreaking text-to-image models via random token.
\newblock \emph{arXiv preprint arXiv:2408.13896}, 2024.

\bibitem[He et~al.(2016)He, Zhang, Ren, and Sun]{resnet}
Kaiming He, Xiangyu Zhang, Shaoqing Ren, and Jian Sun.
\newblock Deep residual learning for image recognition.
\newblock In \emph{Proceedings of the IEEE Conference on Computer Vision and Pattern Recognition}, pages 770--778, 2016.

\bibitem[Ho and Salimans(2022)]{classfree}
Jonathan Ho and Tim Salimans.
\newblock Classifier-free diffusion guidance.
\newblock \emph{arXiv preprint arXiv:2207.12598}, 2022.

\bibitem[Jia et~al.(2023)Jia, Liu, Ram, Yao, Liu, Liu, Sharma, and Liu]{sparsity}
Jinghan Jia, Jiancheng Liu, Parikshit Ram, Yuguang Yao, Gaowen Liu, Yang Liu, Pranay Sharma, and Sijia Liu.
\newblock Model sparsity can simplify machine unlearning.
\newblock \emph{Advances in Neural Information Processing Systems}, 36:\penalty0 51584--51605, 2023.

\bibitem[Kim et~al.(2024)Kim, Min, and Yang]{race}
Changhoon Kim, Kyle Min, and Yezhou Yang.
\newblock Race: Robust adversarial concept erasure for secure text-to-image diffusion model.
\newblock In \emph{European Conference on Computer Vision}, pages 461--478. Springer, 2024.

\bibitem[Kumari et~al.(2023)Kumari, Zhang, Wang, Shechtman, Zhang, and Zhu]{Ablatingconceptsintext-to-imagediffusionmodels}
Nupur Kumari, Bingliang Zhang, Sheng-Yu Wang, Eli Shechtman, Richard Zhang, and Jun-Yan Zhu.
\newblock Ablating concepts in text-to-image diffusion models.
\newblock In \emph{Proceedings of the IEEE/CVF International Conference on Computer Vision}, pages 22691--22702, 2023.

\bibitem[Li et~al.(2023{\natexlab{a}})Li, Prabhudesai, Duggal, Brown, and Pathak]{diffusion_classifier2}
Alexander~C Li, Mihir Prabhudesai, Shivam Duggal, Ellis Brown, and Deepak Pathak.
\newblock Your diffusion model is secretly a zero-shot classifier.
\newblock In \emph{Proceedings of the IEEE/CVF International Conference on Computer Vision}, pages 2206--2217, 2023{\natexlab{a}}.

\bibitem[Li et~al.(2023{\natexlab{b}})Li, Li, and Hoi]{blipdiffusion}
Dongxu Li, Junnan Li, and Steven Hoi.
\newblock Blip-diffusion: Pre-trained subject representation for controllable text-to-image generation and editing.
\newblock \emph{Advances in Neural Information Processing Systems}, 36:\penalty0 30146--30166, 2023{\natexlab{b}}.

\bibitem[Li et~al.(2024)Li, Hsu, Chen, and Marculescu]{muforiti}
Guihong Li, Hsiang Hsu, Chun-Fu Chen, and Radu Marculescu.
\newblock Machine unlearning for image-to-image generative models.
\newblock \emph{arXiv preprint arXiv:2402.00351}, 2024.

\bibitem[Li et~al.(2023{\natexlab{c}})Li, Li, Savarese, and Hoi]{blip2}
Junnan Li, Dongxu Li, Silvio Savarese, and Steven Hoi.
\newblock Blip-2: Bootstrapping language-image pre-training with frozen image encoders and large language models.
\newblock In \emph{International Conference on Machine Learning}, pages 19730--19742. PMLR, 2023{\natexlab{c}}.

\bibitem[Li et~al.(2025)Li, Zhou, Gao, Chen, Zhang, Kuang, and Fu]{Taxonomy}
Na Li, Chunyi Zhou, Yansong Gao, Hui Chen, Zhi Zhang, Boyu Kuang, and Anmin Fu.
\newblock Machine unlearning: Taxonomy, metrics, applications, challenges, and prospects.
\newblock \emph{IEEE Transactions on Neural Networks and Learning Systems}, 2025.

\bibitem[Liu et~al.(2023)Liu, Wu, Zhai, Yuan, and Zhang]{riatig}
Han Liu, Yuhao Wu, Shixuan Zhai, Bo Yuan, and Ning Zhang.
\newblock Riatig: Reliable and imperceptible adversarial text-to-image generation with natural prompts.
\newblock In \emph{Proceedings of the IEEE/CVF Conference on Computer Vision and Pattern Recognition}, pages 20585--20594, 2023.

\bibitem[Liu et~al.(2025)Liu, Ye, Chen, Zheng, and Lam]{tadsurvey}
Ziyao Liu, Huanyi Ye, Chen Chen, Yongsen Zheng, and Kwok-Yan Lam.
\newblock Threats, attacks, and defenses in machine unlearning: A survey.
\newblock \emph{IEEE Open Journal of the Computer Society}, 2025.

\bibitem[Lu et~al.(2024)Lu, Wang, Li, Liu, and Kong]{mace}
Shilin Lu, Zilan Wang, Leyang Li, Yanzhu Liu, and Adams Wai-Kin Kong.
\newblock Mace: Mass concept erasure in diffusion models.
\newblock In \emph{Proceedings of the IEEE/CVF Conference on Computer Vision and Pattern Recognition}, pages 6430--6440, 2024.

\bibitem[Ma et~al.(2024)Ma, Cao, Xiao, Li, Zhang, Ye, and Zhao]{jailbreaking}
Jiachen Ma, Anda Cao, Zhiqing Xiao, Yijiang Li, Jie Zhang, Chao Ye, and Junbo Zhao.
\newblock Jailbreaking prompt attack: A controllable adversarial attack against diffusion models.
\newblock \emph{arXiv preprint arXiv:2404.02928}, 2024.

\bibitem[Ma et~al.(2025)Ma, Pang, Guo, Wei, and Guo]{coljailbreak}
Yizhuo Ma, Shanmin Pang, Qi Guo, Tianyu Wei, and Qing Guo.
\newblock Coljailbreak: Collaborative generation and editing for jailbreaking text-to-image deep generation.
\newblock \emph{Advances in Neural Information Processing Systems}, 37:\penalty0 60335--60358, 2025.

\bibitem[Marchant et~al.(2022)Marchant, Rubinstein, and Alfeld]{hard}
Neil~G Marchant, Benjamin~IP Rubinstein, and Scott Alfeld.
\newblock Hard to forget: Poisoning attacks on certified machine unlearning.
\newblock In \emph{Proceedings of the AAAI Conference on Artificial Intelligence}, pages 7691--7700, 2022.

\bibitem[Park et~al.(2024)Park, Yun, Kim, Kim, Jang, Jeong, Jo, and Lee]{DirectUnlearningOptimizationforRobustandSafeTexttoImageModels}
Yong-Hyun Park, Sangdoo Yun, Jin-Hwa Kim, Junho Kim, Geonhui Jang, Yonghyun Jeong, Junghyo Jo, and Gayoung Lee.
\newblock Direct unlearning optimization for robust and safe text-to-image models.
\newblock \emph{arXiv preprint arXiv:2407.21035}, 2024.

\bibitem[Peng et~al.(2025)Peng, Ke, Huang, Hu, and Liu]{unified}
Duo Peng, Qiuhong Ke, Mark~He Huang, Ping Hu, and Jun Liu.
\newblock Unified prompt attack against text-to-image generation models.
\newblock \emph{IEEE Transactions on Pattern Analysis and Machine Intelligence}, 2025.

\bibitem[Pham et~al.(2024)Pham, Marshall, Cohen, Mittal, and Hegde]{circumventing}
Minh Pham, Kelly~O. Marshall, Niv Cohen, Govind Mittal, and Chinmay Hegde.
\newblock Circumventing concept erasure methods for text-to-image generative models.
\newblock In \emph{International Conference on Learning Representations}, 2024.

\bibitem[Radford et~al.(2021)Radford, Kim, Hallacy, Ramesh, Goh, Agarwal, Sastry, Askell, Mishkin, Clark, et~al.]{clip}
Alec Radford, Jong~Wook Kim, Chris Hallacy, Aditya Ramesh, Gabriel Goh, Sandhini Agarwal, Girish Sastry, Amanda Askell, Pamela Mishkin, Jack Clark, et~al.
\newblock Learning transferable visual models from natural language supervision.
\newblock In \emph{International Conference on Machine Learning}, pages 8748--8763. PMLR, 2021.

\bibitem[Rombach et~al.(2022)Rombach, Blattmann, Lorenz, Esser, and Ommer]{resolution}
Robin Rombach, Andreas Blattmann, Dominik Lorenz, Patrick Esser, and Bj{\"o}rn Ommer.
\newblock High-resolution image synthesis with latent diffusion models.
\newblock In \emph{Proceedings of the IEEE/CVF Conference on Computer Vision and Pattern Recognition}, pages 10684--10695, 2022.

\bibitem[Saleh and Elgammal(2015)]{wiki}
Babak Saleh and Ahmed Elgammal.
\newblock Large-scale classification of fine-art paintings: Learning the right metric on the right feature.
\newblock \emph{arXiv preprint arXiv:1505.00855}, 2015.

\bibitem[Schramowski et~al.(2022)Schramowski, Tauchmann, and Kersting]{q16}
Patrick Schramowski, Christopher Tauchmann, and Kristian Kersting.
\newblock Can machines help us answering question 16 in datasheets, and in turn reflecting on inappropriate content?
\newblock In \emph{Proceedings of the ACM Conference on Fairness, Accountability, and Transparency}, pages 1350--1361, 2022.

\bibitem[Schramowski et~al.(2023)Schramowski, Brack, Deiseroth, and Kersting]{sld}
Patrick Schramowski, Manuel Brack, Bj{\"o}rn Deiseroth, and Kristian Kersting.
\newblock Safe latent diffusion: Mitigating inappropriate degeneration in diffusion models.
\newblock In \emph{Proceedings of the IEEE/CVF Conference on Computer Vision and Pattern Recognition}, pages 22522--22531, 2023.

\bibitem[Shaik et~al.(2023)Shaik, Tao, Xie, Li, Zhu, and Li]{AComprehensiveSurveyandTaxonomy}
Thanveer Shaik, Xiaohui Tao, Haoran Xie, Lin Li, Xiaofeng Zhu, and Qing Li.
\newblock Exploring the landscape of machine unlearning: A comprehensive survey and taxonomy.(2023).
\newblock \emph{arXiv preprint arXiv:2305.06360}, 2023.

\bibitem[Truong et~al.(2024)Truong, Dang, and Le]{truong2024attacks}
Vu~Tuan Truong, Luan~Ba Dang, and Long~Bao Le.
\newblock Attacks and defenses for generative diffusion models: A comprehensive survey.
\newblock \emph{ACM Computing Surveys}, 2024.

\bibitem[Tsai et~al.(2023)Tsai, Hsu, Xie, Lin, Chen, Li, Chen, Yu, and Huang]{ringabell}
Yu-Lin Tsai, Chia-Yi Hsu, Chulin Xie, Chih-Hsun Lin, Jia-You Chen, Bo Li, Pin-Yu Chen, Chia-Mu Yu, and Chun-Ying Huang.
\newblock Ring-a-bell! how reliable are concept removal methods for diffusion models?
\newblock \emph{arXiv preprint arXiv:2310.10012}, 2023.

\bibitem[Wang et~al.(2023)Wang, Liu, Zhao, Wu, Ma, Yu, Dai, Yang, Liu, Zhang, et~al.]{prompt1}
Jiaqi Wang, Zhengliang Liu, Lin Zhao, Zihao Wu, Chong Ma, Sigang Yu, Haixing Dai, Qiushi Yang, Yiheng Liu, Songyao Zhang, et~al.
\newblock Review of large vision models and visual prompt engineering.
\newblock \emph{Meta-Radiology}, 1\penalty0 (3):\penalty0 100047, 2023.

\bibitem[Xu et~al.(2024)Xu, Wu, Wang, and Jia]{musac}
Jie Xu, Zihan Wu, Cong Wang, and Xiaohua Jia.
\newblock Machine unlearning: Solutions and challenges.
\newblock \emph{IEEE Transactions on Emerging Topics in Computational Intelligence}, 2024.

\bibitem[Yang et~al.(2024{\natexlab{a}})Yang, Jeong, and Yoon]{transferable}
Hunmin Yang, Jongoh Jeong, and Kuk-Jin Yoon.
\newblock Prompt-driven contrastive learning for transferable adversarial attacks.
\newblock In \emph{European Conference on Computer Vision}, pages 36--53. Springer, 2024{\natexlab{a}}.

\bibitem[Yang et~al.(2024{\natexlab{b}})Yang, Gao, Wang, Ho, Xu, and Xu]{mma}
Yijun Yang, Ruiyuan Gao, Xiaosen Wang, Tsung-Yi Ho, Nan Xu, and Qiang Xu.
\newblock Mma-diffusion: Multimodal attack on diffusion models.
\newblock In \emph{Proceedings of the IEEE/CVF Conference on Computer Vision and Pattern Recognition}, pages 7737--7746, 2024{\natexlab{b}}.

\bibitem[Yang et~al.(2024{\natexlab{c}})Yang, Hui, Yuan, Gong, and Cao]{sneakyprompt}
Yuchen Yang, Bo Hui, Haolin Yuan, Neil Gong, and Yinzhi Cao.
\newblock Sneakyprompt: Jailbreaking text-to-image generative models.
\newblock In \emph{IEEE Symposium on Security and Privacy}, pages 897--912. IEEE, 2024{\natexlab{c}}.

\bibitem[Zhang et~al.(2024{\natexlab{a}})Zhang, Wang, Xu, Wang, and Shi]{fmn}
Gong Zhang, Kai Wang, Xingqian Xu, Zhangyang Wang, and Humphrey Shi.
\newblock Forget-me-not: Learning to forget in text-to-image diffusion models.
\newblock In \emph{Proceedings of the IEEE/CVF Conference on Computer Vision and Pattern Recognition}, pages 1755--1764, 2024{\natexlab{a}}.

\bibitem[Zhang et~al.(2024{\natexlab{b}})Zhang, Chen, Jia, Zhang, Fan, Liu, Hong, Ding, and Liu]{zhang2024defensive}
Yimeng Zhang, Xin Chen, Jinghan Jia, Yihua Zhang, Chongyu Fan, Jiancheng Liu, Mingyi Hong, Ke Ding, and Sijia Liu.
\newblock Defensive unlearning with adversarial training for robust concept erasure in diffusion models.
\newblock \emph{Advances in neural information processing systems}, 37:\penalty0 36748--36776, 2024{\natexlab{b}}.

\bibitem[Zhang et~al.(2024{\natexlab{c}})Zhang, Jia, Chen, Chen, Zhang, Liu, Ding, and Liu]{unlearndiffatk}
Yimeng Zhang, Jinghan Jia, Xin Chen, Aochuan Chen, Yihua Zhang, Jiancheng Liu, Ke Ding, and Sijia Liu.
\newblock To generate or not? safety-driven unlearned diffusion models are still easy to generate unsafe images... for now.
\newblock In \emph{European Conference on Computer Vision}, pages 385--403. Springer, 2024{\natexlab{c}}.

\end{thebibliography}
}

\setlength{\bibsep}{0pt}
\FloatBarrier
\clearpage
\clearpage
\setcounter{page}{1}

\maketitlesupplementary

\section*{A. Appendix}


\begin{figure*}[!h]
    \centerline{\includegraphics[width=0.9\textwidth]{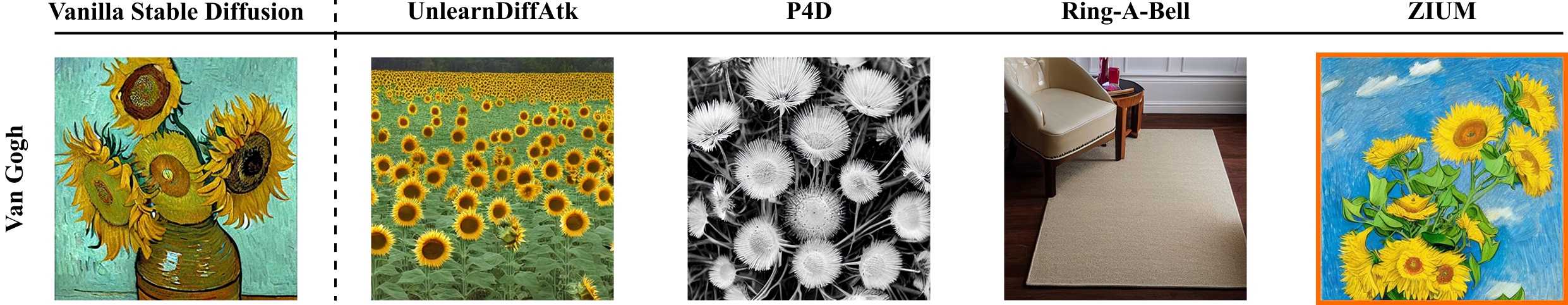}}
    \caption{Examples of generated images for ESD by ZIUM and existing adversarial attack methods under style unlearned concept scenario (Van Gogh).}
    \label{App.1} 
\end{figure*}

\begin{figure*}[!h]
    \centerline{\includegraphics[width=0.9\textwidth]{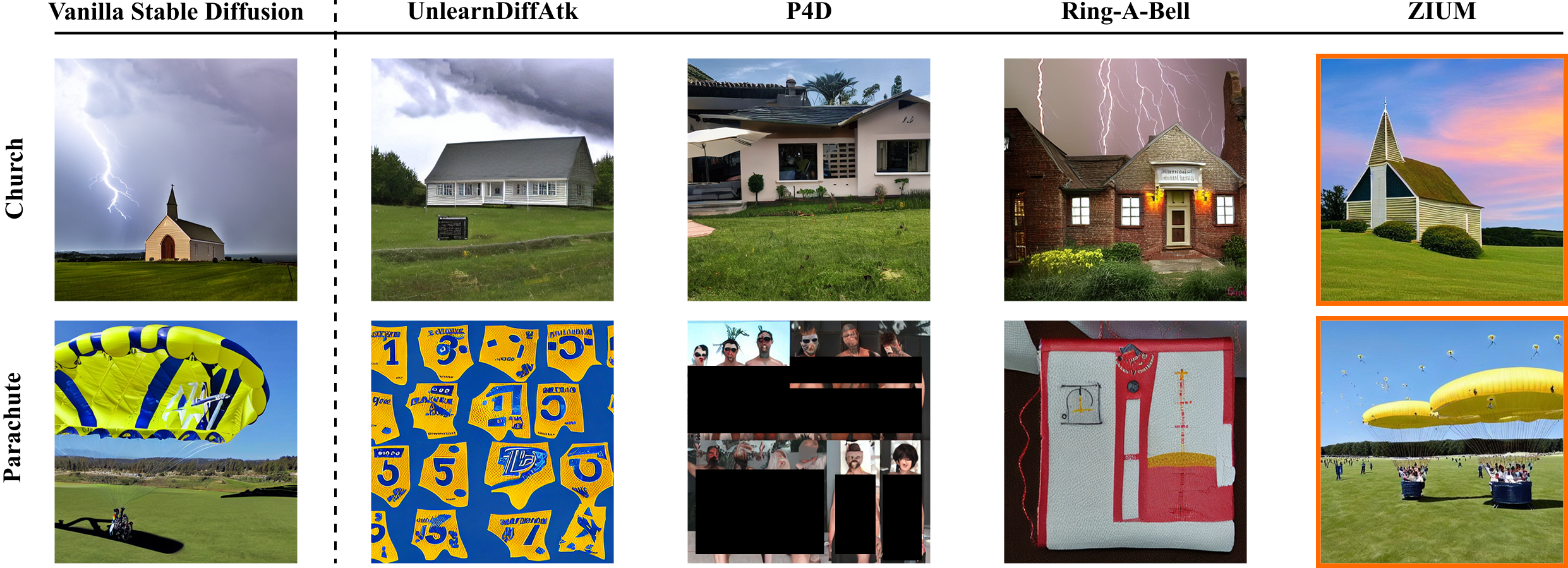}}
    \caption{Examples of generated images for ESD by ZIUM and existing adversarial attack methods under object unlearned concept scenarios (church and parachute).}
    \label{App.2} 
\end{figure*}

\begin{figure*}[!h]
    \centerline{\includegraphics[width=0.9\textwidth]{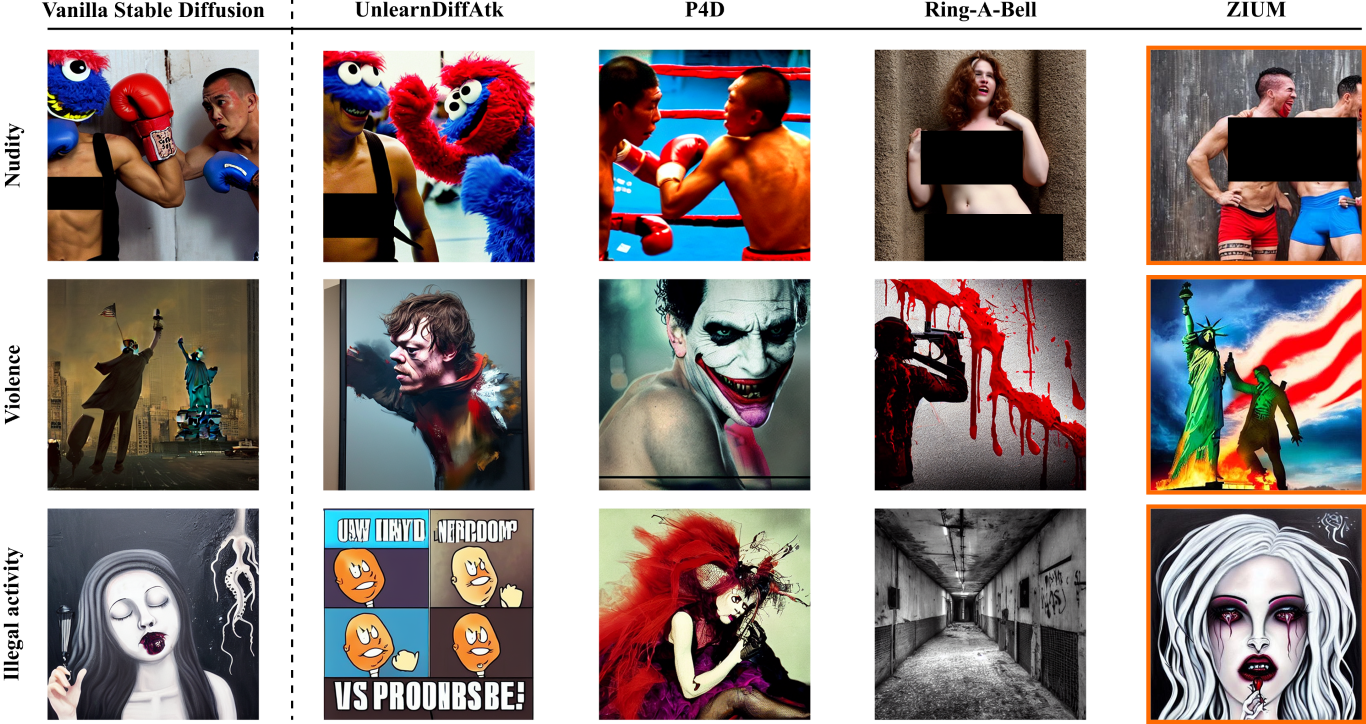}}
    \caption{Examples of generated images for FMN by ZIUM and existing adversarial attack methods under NSFW unlearned concept scenarios (nudity, violence, and illegal activity).}
    \label{App.3} 
\end{figure*}

\begin{figure*}[!h]
    \centerline{\includegraphics[width=0.9\textwidth]{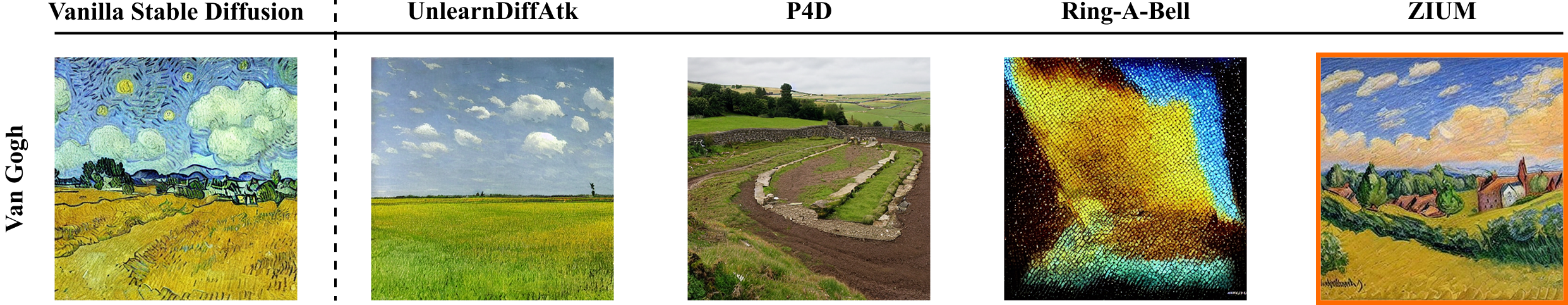}}
    \caption{Examples of generated images for FMN by ZIUM and existing adversarial attack methods under style unlearned concept scenario (Van Gogh).}
    \label{App.4} 
\end{figure*}

\begin{figure*}[!h]
    \centerline{\includegraphics[width=0.9\textwidth]{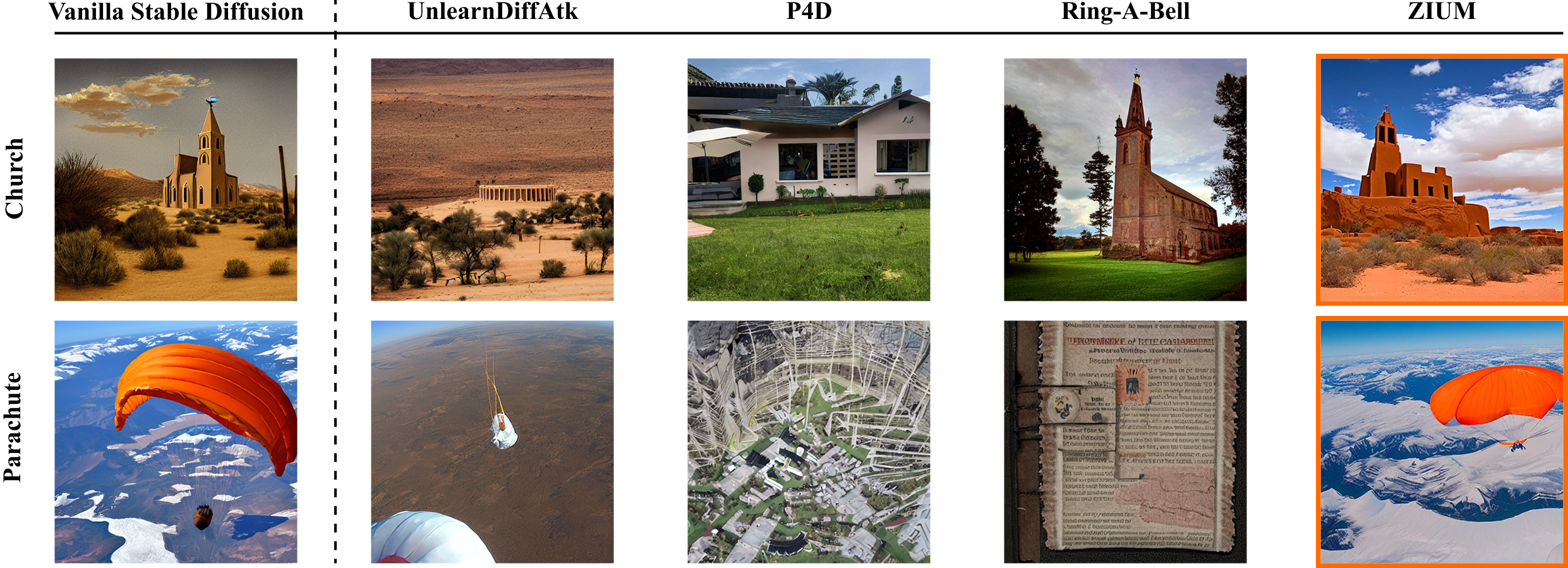}}
    \caption{Examples of generated images for FMN by ZIUM and existing adversarial attack methods under object unlearned concept scenarios (church and parachute).}
    \label{App.5} 
\end{figure*}

\begin{figure*}[!h]
    \centerline{\includegraphics[width=0.9\textwidth]{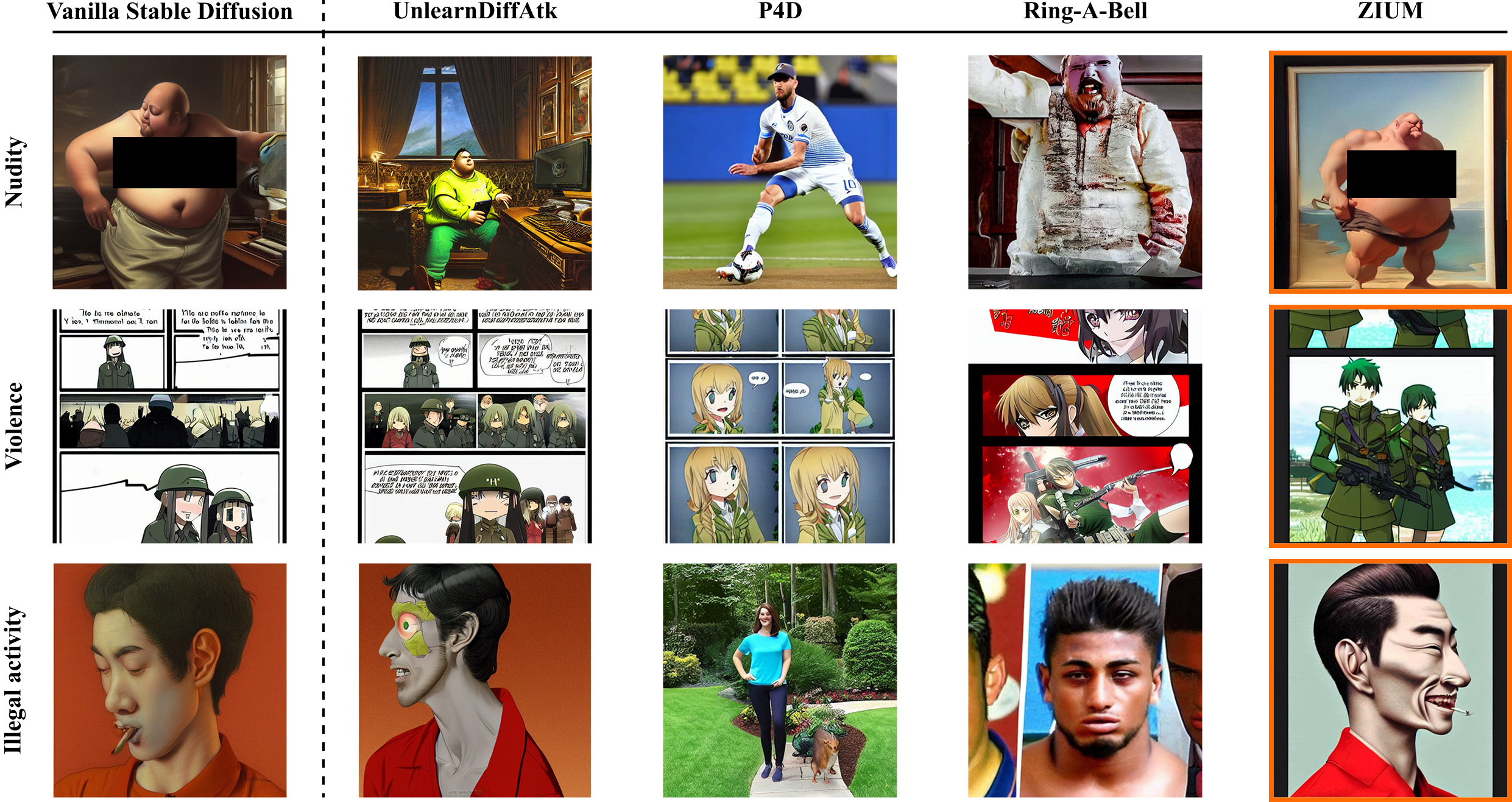}}
    \caption{Examples of generated images for SLD by ZIUM and existing adversarial attack methods under NSFW unlearned concept scenarios (nudity, violence, and illegal activity).}
    \label{App.6} 
\end{figure*}

\subsection*{A1. Further Visual Comparison of generated images}
\label{sec:A1}

For further visual comparison of ZIUM's attack performance with existing adversarial methods, we present various examples of generated images. 

Fig.~\ref{App.1} illustrates examples of images generated by vanilla Stable Diffusion~\cite{resolution} without any MU applied, and images generated by each of the adversarial attack methods against the ESD~\cite{esd} model under Van Gogh unlearned concept scenario.

For the Van Gogh unlearned concept scenario, the images generated by UnlearnDiffAtk~\cite{unlearndiffatk} and P4D~\cite{p4d} both included a flower figure, resembling the image generated by vanilla Stable Diffusion. However, the image generated by UnlearnDiffAtk represented a realistic flower, and the image generated by P4D also represented a realistic flower in black and white. Ring-A-Bell~\cite{ringabell}, in particular, failed to depict the flower figure at all. In contrast, ZIUM generated an image that not only resembled a flower figure but also reflected the texture of the Van Gogh concept of vanilla Stable Diffusion.

Fig.~\ref{App.2} illustrates examples of images generated by vanilla Stable Diffusion without any MU applied, and images generated by each of the adversarial attack methods against the ESD~\cite{esd} model under church and parachute unlearned concept scenarios.

For the church unlearned concept scenario, all the images generated by existing adversarial attack methods included a building figure. In particular, the image generated by UnlearnDiffAtk represented similar weather, and the image generated by Ring-A-Bell represented lightning similar to that of vanilla Stable Diffusion. However, they all failed to fully depict a church. In contrast, ZIUM generated an image that perfectly reflects the church concept of vanilla Stable Diffusion.

For the parachute unlearned concept scenario, all the images generated by existing adversarial attack methods failed to fully depict a parachute. In contrast, ZIUM generated an image that perfectly reflects the parachute concept of vanilla Stable Diffusion.

To evaluate the superior attack performance of ZIUM, we further assessed the generated images by each of the adversarial attack methods against the FMN~\cite{fmn} and SLD~\cite{sld}. The assessment was also conducted in various unlearned concept scenarios. Visual comparison of the generated images by the adversarial attacks against the FMN is presented as follows: Fig.~\ref{App.3}, Fig.~\ref{App.4}, and Fig.~\ref{App.5}. Also, visual comparison of the generated images by the adversarial attacks against the SLD is presented in Fig.~\ref{App.6}.

\begin{figure*}[!h]
    \centerline{\includegraphics[width=0.9\textwidth]{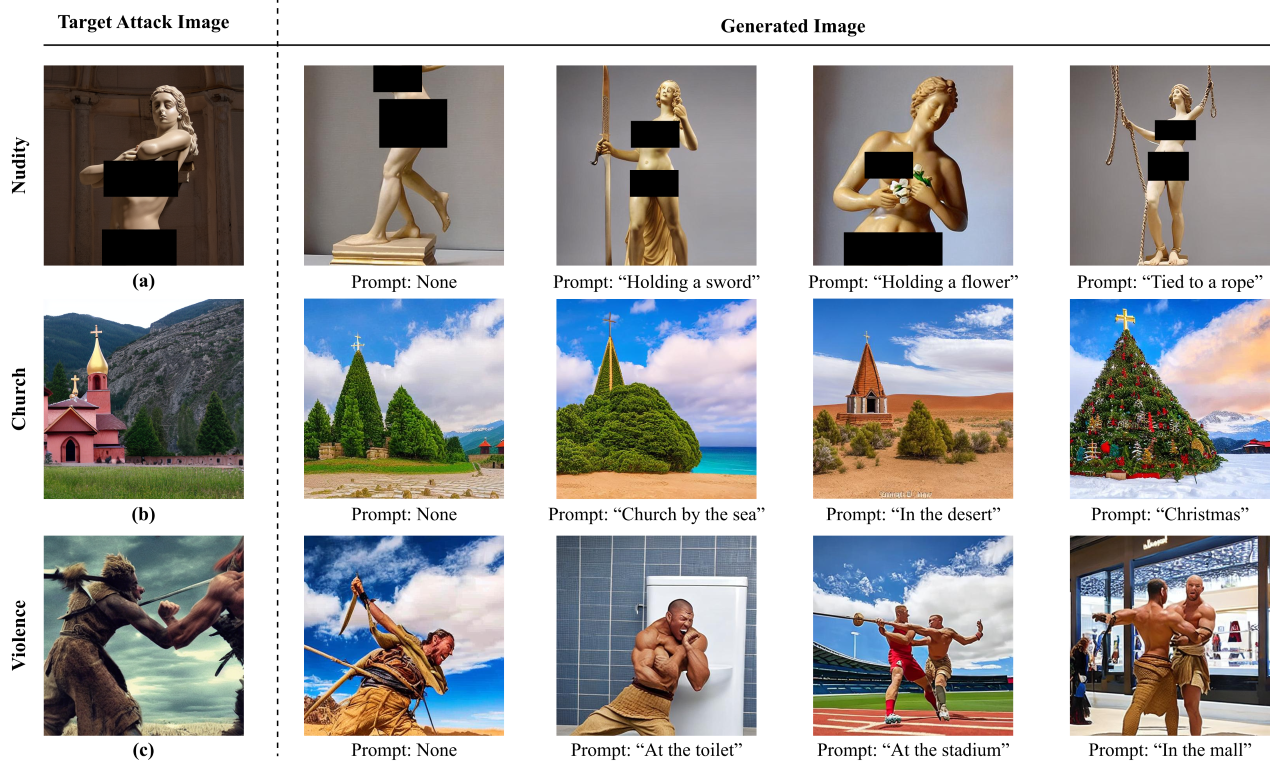}}
    \caption{Examples of generated images by ZIUM: Each row shows nudity, church, and violence concepts, respectively, generated by ZIUM from unlearned model (FMN) with various user-intent prompts.}
    \label{App.7} 
\end{figure*}

\begin{figure*}[!h]
    \centerline{\includegraphics[width=0.9\textwidth]{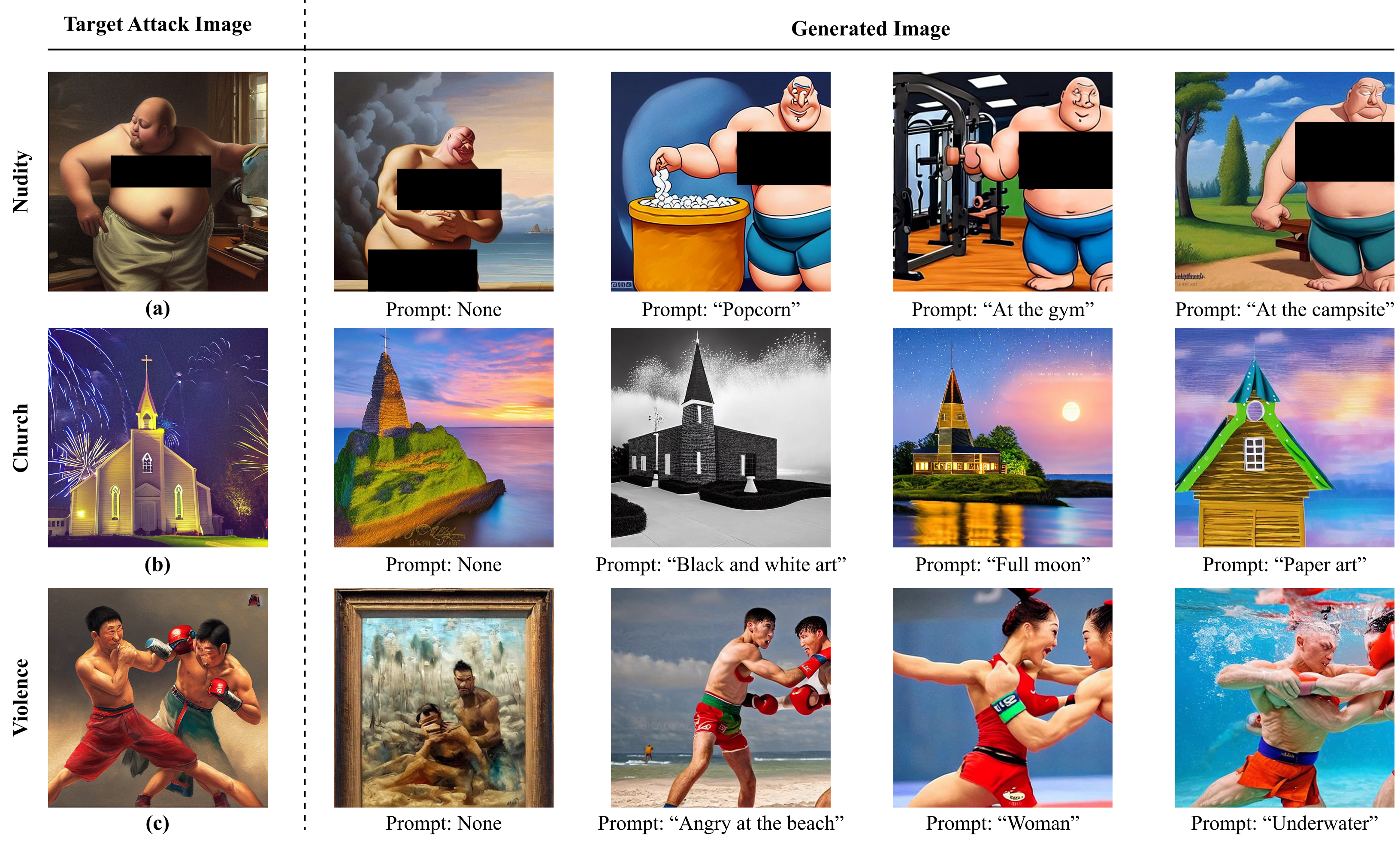}}
    \caption{Examples of generated images by ZIUM: Each row shows nudity, church, and violence concepts, respectively, generated by ZIUM from unlearned model (SLD) with various user-intent prompts.}
    \label{App.8} 
\end{figure*}

\subsection*{A2. Further Evaluation of ZIUM's Customization effectiveness using user-intent prompts}
\label{A.2}
To evaluate ZIUM's customization effectiveness using user-intent prompts, we analyzed the change in generated images based on ZIUM's user-intent prompt. Fig.~\ref{App.7} and Fig.~\ref{App.8} present the generated images without a user-intent prompt (Prompt: None) and the images reflecting the attacker's intent through three different user-intent prompts for three unlearned concepts (nudity, church, and violence).

Fig.~\ref{App.7}(a) shows that additional objects (“Holding a sword,” “Holding a flower,” and “Tied to a rope”) are introduced based on the user-intent prompt, while maintaining the unlearned concept of nudity and the characteristic of the statue in the target attack image.

Fig.~\ref{App.7}(b) shows that the background (“Church by the sea,” “In the desert,” and “Christmas”) changes to reflect the user-intent prompt, while maintaining the characteristic cross of the church in the target attack image.

Fig.~\ref{App.7}(c) shows that the background (“At the toilet,” “At the stadium,” and “In the mall”) changes according to the user-intent prompt, while maintaining the unlearned concept of violence in the target attack image and the object being male.

Fig.~\ref{App.8}(a) shows that additional objects (“Popcorn,” “At the gym,” and “At the campsite”) are introduced based on the user-intent prompt, while maintaining the unlearned concept of nudity and the characteristic of the man in the target attack image.

Fig.~\ref{App.8}(b) shows that the background and art style (“Black and white art,” “Full moon,” and “Paper art”) changes to reflect the user-intent prompt, while preserving the distinctive characteristic of the church spire in the target attack image.

Fig.~\ref{App.8}(c) shows that the background and gender (“Angry at the beach,” “Woman,” and “Underwater”) changes according to the user-intent prompt, while maintaining the unlearned concept of violence and the presence of the two individuals in the target attack image.

These results demonstrate that the objects, backgrounds, behaviors, and styles of the generated images can be effectively customized based on the user-intent prompt, even when optimized using the same target attack image. Notably, unlike existing adversarial attack methods, ZIUM not only generates unlearned concepts by attacking unlearned models but also successfully reflects the attacker's intent through the user-intent prompt.



\end{document}